# Navigating with Graph Representations for Fast and Scalable Decoding of Neural Language Models

**Minjia Zhang**    **Xiaodong Liu**    **Wenhan Wang**    **Jianfeng Gao**    **Yuxiong He**
Microsoft
{minjiaz,xiaodl,wenhanw,jfgao,yuxhe}@microsoft.com

## Abstract

Neural language models (NLMs) have recently gained a renewed interest by achieving state-of-the-art performance across many natural language processing (NLP) tasks. However, NLMs are very computationally demanding largely due to the computational cost of the softmax layer over a large vocabulary. We observe that, in decoding of many NLP tasks, only the probabilities of the top-$K$ hypotheses need to be calculated preciously and $K$ is often much smaller than the vocabulary size. This paper proposes a novel softmax layer approximation algorithm, called **F**ast **G**raph **D**ecoder (FGD), which quickly identifies, for a given context, a set of $K$ words that are most likely to occur according to a NLM. We demonstrate that FGD reduces the decoding time by an order of magnitude while attaining close to the full softmax baseline accuracy on neural machine translation and language modeling tasks. We also prove the theoretical guarantee on the softmax approximation quality.

## 1 Introduction

Drawing inspiration from biology and neurophysiology, recent progress for many natural language processing (NLP) tasks has been remarkable with deep neural network based approaches, including machine translation [1–3], sentence summarization [4], dialogue agents [5], speech recognition [6–8], and conversational bots [9, 10]. Such approaches often employ a neural language model (NLM) as a decoder at inference time to generate a sequence of tokens (e.g., words) given an input, typically via beam search [1–13].

One long-recognized issue of decoding using NLMs is the computational complexity, which easily becomes a bottleneck when the vocabulary size is large. Consider a beam search decoder using a NLM. At each decoding step, a recurrent neural network [14, 15] first generates a context vector based on each partial hypothesis in the beam. It then uses a softmax layer to compute a normalized word probability distribution over the vocabulary [16–18]. The softmax layer consists of an inner product operator that projects the context vector into a vocabulary-sized vector of scores, followed by a softmax function that transforms a vocabulary-sized logits into a vector of probabilities. Finally, the decoder selects top-$K$ words with the highest probabilities given the context (i.e., *top-$K$ maximum subset* of inner product), and stores the expended hypotheses and their probabilities in the beam. The most computationally expensive part in this process is the softmax layer, where the complexity is linear with respect to the vocabulary size. In this paper we strive to develop new softmax approximation methods for fast decoding.

Many techniques have been proposed to speed up the softmax layer in training, such as hierarchical softmax [19] and sampling-based approaches [20–23]. However, most of them cannot be directly applied to decoding because they rely on knowing the words to be predicted and need to calculate the probability of all words to find the most likely prediction during decoding. Other works speed up softmax inference in training and decoding by reducing the cost of computing each word's probability using some approximation [22, 24–26]. However, the complexity of softmax as a whole is still linear with respect to the size of the vocabulary. We observe that in beam-search-based decoding of many NLP tasks we only need to identify the top-$K$ words that are most likely to occur given a context. Can we figure out a search algorithm that can identify these $K$ words without going over the entire large vocabulary? Our answer is yes. Before we present our approach, we review briefly the finding in biological science that motivates our research.

In spite of the large number of words in a vocabulary, human brain is capable of managing them effectively and navigating the massive mental lexicon very efficiently. How is it possible? How is the vocabulary stored and represented in human brain? One of the theories from biological science indicates that human language has a character of complex network [27–29], where the intrinsic relational structure, which refers to the fact that words are related to each other and thus form a *small world graph*, provides some hints on how the lexicon is mentally organized. To predict the next word given a context, humans never need to examine every word in the vocabulary stored in their brains. Instead, a person can immediately identify a small set of $K$ candidate words that are most semantically related to the context, and then pick the most proper word among the candidates. We believe that if we can represent the vocabulary of a NLM using a similar data structure of *small world graph*, we can significantly improve the decoding efficiency of NLMs because at each decoding step softmax only needs to explicitly compute the probabilities of $K$ words, where $K$ is much smaller than the vocabulary size.

We propose a **F**ast **G**raph **D**ecoder (FGD) to approximate the softmax layer of a NLM in the beam search decoding process. First, we construct a *small world graph* representation [30, 31] of a NLM vocabulary. The nodes in the graph are words, each being represented using a continuous vector which is transformed from its word embedding vector in the NLM. The edges in the graph encode the word-word distances in a well-defined metric space. Then, at each decoding step, we identify for a given context (i.e., a partial hypothesis in the beam) the top-$K$ hypotheses and compute their probabilities in the softmax layer of the NLM. We prove that finding the top-$K$ hypotheses in the softmax layer is equivalent to finding the $K$ nearest neighbors using FGD in the small world graph, and the latter can be performed approximately using an efficient graph navigating method [32, 33]. We also prove that the decoding error due to use of the approximated $K$ nearest neighbor search with graph navigation is theoretically bounded.

We validate the effectiveness of our approach on two NLP tasks, neural machine translation and language modeling. Empirical results show that FGD achieves an order of magnitude speedup while attaining the accuracy, in comparison with existing state-of-the-art approaches.

In the rest of the paper, Section 2 details the softmax implementation and the challenge. Section 3 describes FGD and gives theoretical justifications. Section 4 presents experimental results. Conclusions are drawn in Section 5.

## 2 Motivation and Challenge

The softmax layer of a NLM is the computational bottleneck at the decoding time in many NLP tasks. Consider a NLM that uses a two–layer LSTM and a vocabulary size of $|V|$ [16–18]. The total number of floating point operations (FLOPS) per LSTM step is $2(\text{layer}) \times (I + D) \times D \times 4 \times 2$ [1], where $I$ and $D$ represent the input and hidden dimension, respectively. The number of FLOPS of the softmax layer is $D \times |V| \times 2$, which is proportional to $|V|$. Assuming that the dimension of the input/hidden layers of the LSTM is 500 and the vocabulary size is 50K, the LSTM part has 8M FLOPS whereas the softmax layer has 50M FLOPS. The softmax layer dominates the computational cost of the NLM, and even more so with a larger vocabulary.

This decoding bottleneck limits NLMs' application in many interactive services such as Web search, online recommendation systems and conversational bots, where low latency, often at the scale of milliseconds, is demanded. In addition, unlike model training where we leverage massive parallelism power of GPUs, decoding needs to run in various clients ranging from PC, mobile, to IoT (Internet of Things), most of which have limited hardware resources and where GPUs are not always available [34]. Therefore, fast decoding is crucial to broaden the applicability of NLMs.

---

[1] It times 4 because an LSTM has 3 gates and 1 memory cell, and it times 2 because each weight value causes a multiply–and–add operation.

## 3 Approach

We propose **F**ast **G**raph **D**ecoder (FGD) to approximate the softmax layer in NLMs. FGD works in two steps, as illustrated in Figure 1. First, for any given NLM, we construct a small world graph to represent its vocabulary. Second, at each step in beam search decoding, we find for each partial hypothesis in the beam (i.e., context) the top-$K$ words that are most likely to occur, and store the expended hypotheses and their probabilities in the beam.

FGD is expected to be far more efficient than the full softmax in that

- when expanding a partial hypothesis we only need to explicitly compute the probabilities of top-$K$ words instead of every word in a vocabulary, and

- there is an efficient search method of identifying the top-$K$ words on the small world graph without going over the entire vocabulary.

In what follows, we will present in turn

- Why do we choose the small world graph representation in FGD? (Section 3.1)

- How to construct a small world graph to represent the vocabulary of a NLM? (Section 3.2)

- How to identify top-$K$ hypotheses for a given context on small world graph? (Section 3.3)

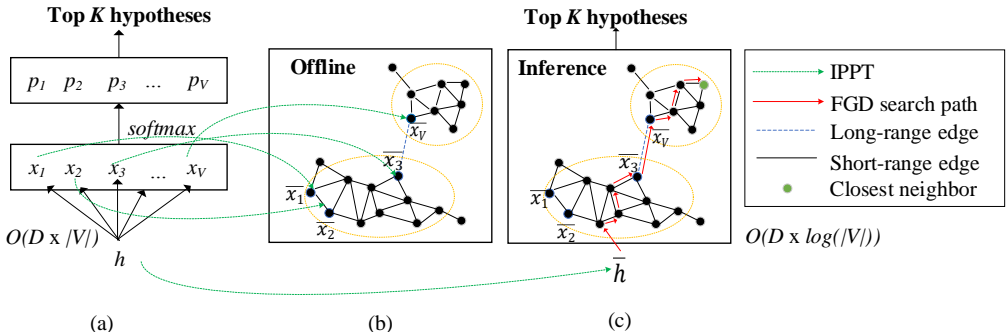

Figure 1: **Overview of FGD:** (a) illustrates one decoding step using a NLM with a vocabulary $V$, where given a context vector $h \in \mathbb{R}^D$ top-$K$ hypotheses are selected among all possible hypotheses generated by $V$ using softmax. The complexity of the softmax layer is $O(D \times |V|)$. (b) shows the transformation from the word embedding vectors of the NLM vocabulary, $x_1, x_2, .., x_{|V|}$, to a small world graph representation which encodes word-word distances in a well-defined metric space. This transformation, incurring once offline, is essential for FGD to perform fast decoding. (c) shows one step decoding using FGD. For a given context vector $h$, FGD identifies top-$K$ hypotheses by traversing the small world graph and produces their probabilities. The complexity of the FGD-approximated softmax layer is $O(D \times \log |V|)$.

### 3.1 Why Small World Graphs?

In FGD, finding top-$K$ words for a given context is implemented by finding $K$ nearest neighbors in a vector space. The small world graph has been recently introduced to address the problem of nearest neighbor search [32, 33]. Research shows that navigation in small world graph exhibits $O(\log N)$ search complexity ($N$ represents the number of nodes in the graph), and performs well in high dimensionality [32, 33, 35].

To get logarithmic nearest neighbor search complexity, the small world graph needs to hold the *small world* properties which we will detail in the following sections, such as great local connectivity (as in a lattice graph) combined with a small graph diameter (as in a random graph) [36], and a well-defined pair-wise *metric* distance among words.

## 3.2 Small World Graph Construction

Denote $G = (\overline{X}, E)$ as a small world graph with the set $\overline{X}$ as graph nodes and $E$ as the graph edges. Given a NLM vocabulary, which is a set of word embedding vectors $X = [x_1, x_2, ..., x_{|V|}], x_i \in \mathbb{R}^D$, the construction of its small world graph $G$ takes two steps.

1. Constructing the node set $\overline{X}$: as illustrated in Figure 1(b), for each word embedding vector $x_i$ in $X$, we apply a transformation called Inner Product Preserving Transformation (IPPT) to obtain $\overline{X} = [\overline{x_1}, \overline{x_2}, ..., \overline{x_{|V|}}], \overline{x_i} \in \mathbb{R}^{D+2}$ that establishes the equivalence of finding top-$K$ maximum subset of inner product in $X$ and searching for top-$K$ nearest neighbors in $\overline{X}$ with a given distance metric $\rho$. (Section 3.2.1)

2. Constructing the edge set $E$: Given $\overline{X}$, we impose the set $E$ based on the distance metric $\rho$ to form a small world graph. (Section 3.2.2)

### 3.2.1 Constructing $\overline{X}$ via Inner Product Preserving Transformation

Using inner product over word embedding vectors to measure word-word distance is deficient because it lacks very basic properties that need to hold for distance (i.e., the inverse of similarity) functions in metric spaces (e.g., Euclidean spaces) – identity of indiscernibles and triangle inequality [37]. For example, under the Euclidean space, two points are the same iff their distance is 0. The inner product of a point $x$ to itself is $\|x\|^2$, but there can be other points whose inner product to $x$ is smaller than $\|x\|^2$. The search process on small world graphs relies on these properties to converge and achieve their efficiency [32].

To create a small world graph with a well-defined metric distance between words, we present a new method called Inner Product Preserving Transformation (IPPT) to convert word embedding vectors to higher dimensional vectors. We establish the equivalence of finding top-$K$ maximum subset of inner product and searching for top-$K$ nearest neighbor with a distance metric in the higher dimension space. We use the notation $\langle \cdot, \cdot \rangle$ for the inner product and $\rho(\cdot, \cdot)$ for the distance in Euclidean space. Thus $\rho(\xi, \eta) = \|\xi - \eta\|_2 = \sqrt{\langle \xi - \eta, \xi - \eta \rangle}$.

In lemma 3.1, we show that it is possible to define particular transformation functions for word embedding vectors and the given context vector, respectively, so that the transformation is *inner product preserving*. We present the proof of lemma 3.1 in Appendix A.

**Lemma 3.1.** *Let $x_i \in \mathbb{R}^D$ and $b_i \in \mathbb{R}$ be the word embedding vector and bias at position $i$ in the softmax layer, respectively, for $1 \leq i \leq |V|$. Choose a constant $U$ such that $U \geq \max_{i \in V} \sqrt{\|x_i\|_2^2 + b_i^2}$. Let $[;]$ represents vector concatenation. Define the transformation function for word embedding vectors $\phi : \{(\xi, \eta) \in \mathbb{R}^D \times \mathbb{R} : \|\xi\|_2^2 + \eta^2 \leq U^2\} \longrightarrow \mathbb{R}^{D+2}$ as $\phi(x, b) = \left[x; b; \sqrt{U^2 - \|x\|_2^2 - b^2}\right]$. Define the transformation of the given context vector $h \in \mathbb{R}^D$ as $\overline{h} = [h; 1; 0] \in \mathbb{R}^{D+2}$. These transformations are inner product preserving in that for any $i \in V$, we have $\langle h, x_i \rangle + b_i = \langle \overline{h}, \phi(x_i, b_i) \rangle = \frac{1}{2}\left(U^2 + 1 + \|h\|_2^2 - \rho(\overline{h}, \phi(x_i, b_i))^2\right)$.*

The above lemma indicates that the transformation is both "order preserved" and "value preserved". The former means that given a context vector, its top-$K$ closest words identified using inner product in the word embedding space are the same as identified according to the Euclidean distance in the transformed space. The latter means that the ratio of the two distance scores between the context and a word computed in the two spaces respectively is a constant. Lemma 3.1 implies that given a context, we can find the same top-$K$ hypotheses in either space. In the context of decoding using a NLM, the lemma implies that given a context, finding its top-$K$ words using the softmax layer in the NLM is equivalent to finding its $K$ nearest neighbors in the small world graph constructed from the NLM vocabulary. This is formally stated in Theorem 3.2.

**Definition 1** (**Top-$K$ maximum (minimum) subset**). *Let $V$ be the set of vocabulary, and $1 \leq K \leq |V|$. We call $\mathcal{K}$ a top-$K$ maximum (minimum) subset for a function $f : V \to \mathbb{R}$, if $|\mathcal{K}| = K$ and $f(v_i) \geq f(v_j) (f(v_i) \leq f(v_j))$ for all $v_i \in \mathcal{K}$ and $v_j \notin \mathcal{K}$.*

**Theorem 3.2.** *Suppose $1 \leq K \leq |V|$ and consider a fixed context vector $h$. Let $\mathcal{K} \subseteq V$ be a top-$K$ maximum subset for $v_i \mapsto \langle h, x_i \rangle + b_i$. Then $\mathcal{K}$ is also a top-$K$ minimum subset for the Euclidean distance $v_i \mapsto \rho(\overline{h}, \phi(x_i, b_i))$.*

### 3.2.2 Small World Graph Construction

Now, we are ready to present the algorithm for small world graph construction. The algorithm, termed as FGD–P (P for Preprocessing), is presented in Algorithm 1.

Given a trained NLM with its vocabulary represented as word embedding vectors $X$, FGD–P first constructs the node set $\overline{X}$ using IPPT as described above (in line 4-9 in Algorithm 1).

---

**Algorithm 1**  Offline preprocessing algorithm FGD–P
---
1: **Input:** Trained weights of the softmax layer $X$, and bias vector b.
2: **Output:** Small world graph G, and $U_{max}$.
3: **Hyperparameter:** Small world graph neighbor degree M.
4: **for all** $i$ **in** $(0..|X|-1)$ **do**
5:     $\tilde{X}_{:i} \leftarrow [X_{:i}; b_i]$       ▷ Word embedding and bias fusion
6: $U_{max} \leftarrow \max_i \|\tilde{X}_{:i}\|_2$
7: **for all** $i$ **in** $0..(|\tilde{W}|-1)$ **do**
8:     $\Delta_i \leftarrow \sqrt{U_{max}^2 - \|\tilde{X}_{:i}\|_2^2}$       ▷ Calculate the normalizer
9:     $\overline{X}_{:i} \leftarrow [\tilde{X}_{:i}; \Delta_i]$
10: $G \leftarrow CreateSwg(\overline{X}, M)$       ▷ Build small world graph

---

Then, FGD–P forms the final graph using $G = CreateSwg(\overline{X}, M)$ by inserting edges among nodes (line 10). We need to ensure that $G$ is constructed in the way that all the small world properties are hold. We thus explore existing algorithms that have devoted to constructing a small world graph. Among the most accomplished algorithms, HNSW (Hierarchical Navigable Small Worlds) has recently attained outstanding speed-accuracy trade-offs [33], and thus employed by us in this study. We briefly describe the main ideas and refer readers to Malkov and Yashunin [33] for more details.

The small world graph is built incrementally by iteratively inserting each word vector $\overline{x_i}$ in $\overline{X}$ as a node in $G$. Each node will generate $M$ (i.e., the neighbor degree) out-going edges. Among those, $M-1$ are *short–range* edges, which connect $\overline{x_i}$ to $M-1$ nearest neighbors according to their pair-wise Euclidean distance to $\overline{x_i}$ (e.g., the edge between $\overline{x_1}$ and $\overline{x_2}$ in Figure 1 (b)). The rest is a *long–range* edge that connects $\overline{x_i}$ to a randomly picked node, which does not necessarily connect two closest nodes but may connect other isolated clusters (e.g., the edge between $\overline{x_3}$ and $\overline{x_{|V|}}$ in Figure 1 (b)). It is theoretically justified that constructing $G$ by inserting these two types of edges guarantees the graph small world properties [32, 33, 36] of the resulting $G$.

The constructed small world graph using all $\overline{x_i}$ becomes the ground layer $L_0$ of $G$. $CreateSwg$ then creates a hierarchical small world graph by selecting a chain of subsets $V = L_0 \supseteq L_1 \supseteq \ldots \supseteq L_l$ of nodes as "layers". This is similar to the HNSW [33]. Each node in $L_k$ is randomly selected to $L_{k+1}$ with a fixed probability $1/M$. On each layer, the edges are defined so that the overall structure becomes a small world graph, and the number of layers is bounded by $O(\log|V|/\log M)$ [32].

### 3.3 Decoding as Searching Small World Graphs

FGD–I (I for Inference) (Algorithm 2) shows how FGD is used for fast decoding (as in Figure 1 (c)). It first transforms the context vector $h$ to $[h; 1; 0] \in \mathbb{R}^{d+2}$ (line 4). Then, it calls $SearchSwg(G, h, K)$ to search in the small world graph to identify the top-$K$ hypotheses using the search method from HNSW [33], which will be briefly described below.

The search of the graph starts from its top layer and uses a greedy search to find the node with the closest distance to $\overline{h}$ as an entry point to descend to the lower layers. The upper layers route $\overline{h}$ to an entry point in the ground layer that is already close to the nearest neighbors to $\overline{h}$. Once reaching the ground layer, $SearchSwvg$ employs prioritized breath-first search: It exams its neighbors and stores all the visited nodes in a priority queue based on their distances to the context vector. The length of the queue is bounded by $efSearch$, a hyperparameter that controls the trade-off between search time and accuracy. When the search reaches a termination condition (e.g., the number of distance calculation), $SearchSwg$ returns the results of top-$K$ hypotheses and their distances to $\overline{h}$. We transform the distance value back to the original inner product space (line 5– 7), as described in Section 3.2.1. FGD–I generates the output by computing a softmax distribution over the inner product of the top-$K$ returned results (line 8).

---

**Algorithm 2**                                                  Online inference algorithm FGD–I

1: **Input:** Context vector $h$, small world graph $G$, and $U_{max}$.
2: **Output:** Probability distribution $P$ over top-$K$ word hypotheses.
3: **Hyperparameter:** Candidate queue length $efSearch$.
4:     $\overline{h} \leftarrow [h; 1; 0]$                                 ▷ Map context vector from $\mathbb{R}^D$ to $\mathbb{R}^{D+2}$
5:     $I^K, D^K \leftarrow SearchSwg(G, \overline{h}, K)$      ▷ Return top-$K$ hypotheses with minimal distance to $\overline{h}$
6: **for all** $i$ **in** $0..(K-1)$ **do**
7:        $S[I_{:i}^K] \leftarrow \frac{1}{2}\left(\|\overline{h}\|_2^2 + U_{max}^2 - D_{:i}^{K\,2}\right)$      ▷ Map Euclidean distance back to inner product
8:     $P \leftarrow exp(S)/\sum exp(S)$                ▷ Compute top-$K$ softmax probability distribution

---

In practice, setting $K = |V|$ is both slow and unnecessary. An approximated approach with $K << |V|$ is often much more efficient without sacrificing much accuracy. We provide a theoretically derived error bound of approximating softmax via computing a probability distribution using only top-$K$ distance scores in Appendix B. In Section 4, we will demonstrate empirically the effectiveness of our approach.

## 4 Evaluation

**Summary of main results.** In this section, we present the results of FGD on two different tasks: neural machine translation (NMT) and language modeling (LM).

1. On NMT, FGD obtains more than 14X speedup on softmax execution time than full-softmax with close to baseline BLUE score, and obtains 30X speedup at the cost of losing 0.67 BLEU score.
2. On LM, FGD scales with a logarithmic increase of execution time and outperforms full-softmax by an order of magnitude with large vocabulary.

**Setup.** We implement FGD in Python using numpy[2]. To construct the small world graph, we employ a state-of-the-art framework NMSLIB [33, 38]. The execution time is given as the averaged per-step decoding time in milliseconds, measured on a 64-bit Linux Ubuntu 16.04 server with two Intel Xeon CPU E5-2650 v4 @ 2.20GHz processor with single thread regime so that all algorithms are compared under the same amount of hardware resource.

### 4.1 Neural Machine Translation

NMT is a sequence–to–sequence model which contains an RNN encoder and an RNN decoder. The decoder contains an output projection at every step to predict the next word. Decoding time and BLEU score [39] are the two major metrics for this evaluation. The lower the decoding time without sacrificing much BLEU score, the better the result. We train a global attention-based [40] encoder–decoder model with a two-unidirectional-stacked LSTM [1, 2] using the OpenNMT-py toolkit [41] on the IWSLT'14 German-English corpus [42]. We set the LSTM hidden dimension size to 200. The model is optimized with SGD using an initial learning rate of 1.0 and a dropout [43] ratio of 0.3. The dataset is tokenized and preprocessed using the OpenNMT data preprocessor with $|V| = 50,000$ frequent words [23, 40]. BLEU score is computed with the Moses toolkit [44].

Once the model is trained, we process the trained weights in the softmax layer using FGD–P offline. It takes three minutes on our server to construct the small world graph. During online processing, the hyperparameter, $efSearch$, decides the length of the candidate queue to track nearest neighbors, which offers the trade-off between the online decoding speed and the BLEU score quality. We test different $efSearch$ values and identify [20, 200] as a good range.

**Decoding time and BLEU score comparison with existing methods.** Two approaches are used for comparison: 1) a full-softmax approach; 2) a state-of-the-art approach, called SVD-softmax [24].

---

[2]http://www.numpy.org/

SVD-softmax improves the inference speed by approximating softmax layer using singular vector decomposition (SVD). It includes two steps: it first estimates the probability of each word using a small part of the softmax weight matrix, and then performs a refinement on top-$\overline{V}$ most likely words based on the previous estimated results. It reduces the complexity from $O(|V| \times D)$ to $O(|V| \times \overline{D} + |\overline{V}| \times D)$, where $1 \leq \overline{D} < D$. As suggested by [24], we use two configurations of SVD-softmax: *SVD-a*[3] and *SVD-b*[4].

Figure 2 shows the main results — FGD achieves significantly lower execution time than the existing methods with comparable BLEU scores.

Comparing with full softmax, when $efSearch$ is 20, FGD reduces the execution time from 6.3ms to 0.21ms, achieving 30X speedup at the cost of losing 0.67 BLEU score. By increasing $efSearch$ to 50, FGD obtains nearly the same BLEU score as the full-softmax baseline, while reducing the execution time from 6.3ms to 0.43ms and achieving more than 14X speedup.

For SVD-softmax, We also observe that *SVD-b* approaches a BLEU score close to the full-softmax baseline, but it is much slower than FGD in terms of the execution time (5.53ms vs 0.43ms). *SVD-a* shows slightly better performance than *SVD-b* but with a lower BLEU score. Although the theoretical speedup of *SVD-a* is 5.5X, it gets only 1.3X speedup in practice because top-$\overline{V}$ most likely words selected in the first step appear at discontinuous location on memory, which causes non-negligible memory copy cost to bring them to a continuous space for the second step calculation.

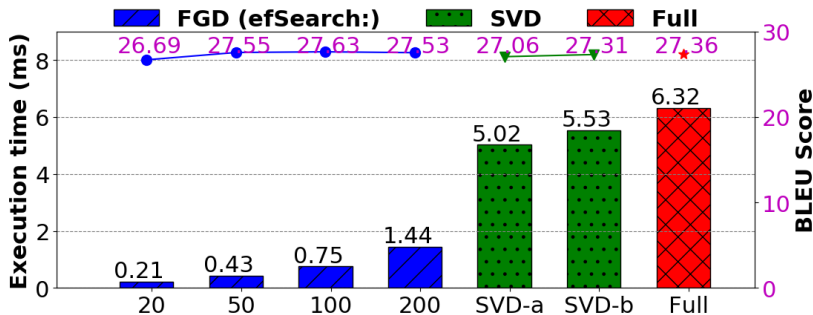

Figure 2: Execution time of the softmax layer and BLEU score of NMT model with FGD, SVD-softmax (**SVD**), and Full-softmax (**Full**). [20, 50, 100, 200] are the hyperparameter of $efSearch$ in FGD. Execution time is displayed as the height of the bar chart, in millisecond (lower is better). BLEU scores are labeled with colored numbers on the top (higher is better).

**Sensitivity of sequence lengths.** Figure 3 reports the results with $efSearch = 100$. FGD is on a par with the full softmax baseline uniformly on different lengths (without statistically significant difference). It demonstrates the robustness of the proposed approach.

**Sensitivity of beam sizes.** We vary the beam size among 1, 2, 5, 10, which are typical settings used by prior work [1–3, 45]. Table 1 shows that, when $efSearch$ is equal or larger than 50, FGD obtains the BLEU scores close to the full softmax baseline under all beam sizes without statistical significance.

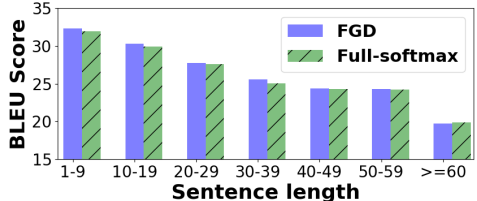

| efSearch | Beam = 1 | Beam = 2 | Beam = 5 | Beam = 10 |
|---|---|---|---|---|
| 20 | 26.69 | 27.65 | 27.81 | 27.62 |
| 50 | 27.55 | 28.76 | 29.06 | 28.9 |
| 100 | 27.63 | 28.94 | 29.28 | 29.1 |
| 200 | 27.53 | 28.99 | 29.28 | 29.22 |
| Full | 27.36 | 28.91 | 29.45 | 29.34 |

Figure 3: BLEU score breakdown by sentence length (setting $efSearch$=100).

Table 1: BLEU score on NMT task, with various beam sizes.

**Internals of FGD.** To reveal the internals of FGD, we analyze two metrics, *precision@K* (or equivalently *P@K*) and *dist_cnt*. *Precision@K* measures the proportion of overlap between retrieved

---
[3]The preview window width $\overline{D}$ is set to 16, and the refinement window width $\overline{V}$ is set to 2500.
[4]The preview window width $\overline{D}$ is set to 16, and the refinement window width $\overline{V}$ is set to 5000.

top-$K$ hypotheses and expected top-$K$ hypotheses, based on what top-$K$ on a full-softmax would return. $dist\_cnt$ measures the number of distance computation in FGD under a given $efSearch$. Table 2 reports *precision@K* when $K$ is 1, 2, 5, and 10, which correspond to beam size 1, 2, 5, and 10 respectively, and $dist\_cnt$ with vs. without FGD. Overall, FGD achieves fairly high precision. In particular, gradually increasing $efSearch$ leads to higher precision at the expense of increased number of distance computation. This matches the observation that higher $efSearch$ leads to higher BLEU score (Figure 2) and also longer execution time (Table 1). Further increasing $efSearch$ leads to little extra precision improvement but significantly more distance computation because the precision is getting close to 1, which explains why FGD can get close to baseline BLEU score (Table 1). We also observe that under the same $efSearch$, further increasing $K$ sometimes leads to slightly worse precision if $efSearch$ is not big enough (e.g., $efSearch$ is 20), as the highest ranked words not visited during the graph search are definitely lost. On the other hand, the computation of distance grows proportional to the increase of $efSearch$. Comparing with the full-softmax, the amount of distance computation is reduced by 10–50 times, which explains the speedup of decoding time (Figure 2).

| efSearch | P@1 | P@2 | P@5 | P@10 | dist_cnt (FGD/ Full) |
|---|---|---|---|---|---|
| 20 | 0.939 | 0.934 | 0.929 | 0.918 | 981 / 50K |
| 50 | 0.974 | 0.974 | 0.973 | 0.971 | 1922 / 50K |
| 100 | 0.986 | 0.986 | 0.987 | 0.987 | 3310 / 50K |
| 200 | 0.992 | 0.993 | 0.994 | 0.994 | 5785 / 50K |

Table 2: Precision and distance computation results on the NMT model.

## 4.2 Language Modeling

This section evaluates the impact of vocabulary size and word embedding dimension on FGD using language modeling [5] on WikiText-2 [46]. The model uses a two–layer LSTM[6].

**Impact of vocabulary size.** We explore multiple models with different vocabulary size of 10,000 (10K), 20,000 (20K), 40,000 (40K), and 80,000 (80K). The vocabulary is created by tokenizing raw texts via Moses toolkit [44] and choosing the correspondingly topmost frequent words on the raw WikiText-2 dataset [46]. Both input and hidden dimension are set to 256.

Table 3 shows the impact of search quality by varying vocabulary size from 10K to 80K. With the same $efSearch$, FGD generally obtains better precision results for smaller vocabulary; With the same vocabulary size, bigger $efSearch$ is better for high precision. With $efSearch$ being 200, FGD is getting very close to 1.

| |V| | P@K | FGD (efSearch) | | | |
|---|---|---|---|---|---|
| | | 20 | 50 | 100 | 200 |
| 10K | P@1 | 0.870 | 0.938 | 0.989 | 1.000 |
| | P@10 | 0.909 | 0.972 | 0.992 | 0.998 |
| 20K | P@1 | 0.845 | 0.932 | 0.975 | 0.995 |
| | P@10 | 0.871 | 0.955 | 0.987 | 0.997 |
| 40K | P@1 | 0.808 | 0.912 | 0.936 | 0.980 |
| | P@10 | 0.845 | 0.931 | 0.961 | 0.991 |
| 80K | P@1 | 0.832 | 0.933 | 0.966 | 0.982 |
| | P@10 | 0.858 | 0.945 | 0.978 | 0.994 |

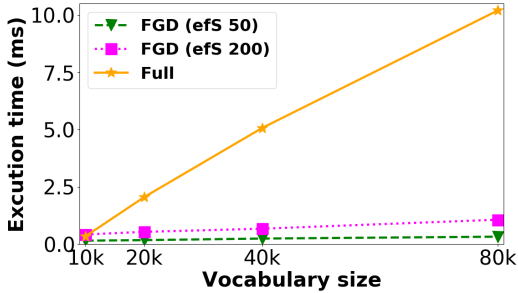

Table 3: Precision of FGD on WikiText-2 dataset varying vocabulary size.

Figure 4: Scalability of WikiText-2 language model varying vocabulary size.

Figure 4 shows the decoding time of varying vocabulary sizes on the full softmax baseline and FGD (settings $efSearch$={50, 200} for the sake of readability). As expected, the execution time all increases with the increase in vocabulary size. However, compared to the baseline, FGD provides a shorter execution time consistently. As the vocabulary size increases, the execution time of the baseline increases almost linearly, whereas FGD's execution time increases much more slowly. This is because the complexity of softmax is $O(D \times |V|)$, which is linear to the size of the vocabulary, whereas the complexity of FGD is $O(D \times \log |V|)$, which is logarithmic to $|V|$. Therefore, FGD scales much better and the improvement becomes more significant with larger vocabulary sizes. In particular, FGD is more than an order of magnitude faster than the baseline when the vocabulary size is medium or large. For example, FGD achieves more than 30X speedup with $|V| = 80K$ when $efSearch = 50$ (Appendix 5 includes a speedup graph).

---
[5] https://github.com/pytorch/examples/tree/master/word_language_model
[6] The models are trained with stochastic gradient descent (SGD) with an initial learning rate of 20 [47]. The batch size is set to 20, and the network is unrolled for 35 timesteps. Dropout is applied to LSTM layers with a dropout ratio of 0.3 [43]. Gradient clipping is set to 0.25 [48].

**Sensitivity of word embedding dimension.** We also test various word embedding dimensions. FGD gets high precision consistently with an order of magnitude execution time reduction (see Appendix C).

## 5 Conclusion

We propose a novel softmax layer approximation algorithm, called **F**ast **G**raph **D**ecoder (FGD), which quickly navigates, for a given context, on a *small world graph* representation of word embeddings to search for a set of $K$ words that are most likely to be the next words according to NLMs. On neural machine translation and neural language modeling, we demonstrate that FGD reduces the decoding time by an order of magnitude (e.g., 14X speedup comparing with the full softmax baseline) while attaining similar accuracy on neural machine translation and language modeling tasks. As the further work, we also like to explore how to speed up NLMs training with large vocabularies.

## A Proof of Lemma and Theorem

*Lemma 3.1 proof.* For the first equation can be proved by expanding the inner product according to the definition of the transformation. Note that $\langle \tilde{h}, \phi(x_i, b_i) \rangle = \left\langle [h; 1; 0], \left[x_i; b_i; \sqrt{U^2 - \|x_i\|^2 - b_i^2}\right] \right\rangle = \langle h, x_i \rangle + 1 \cdot b_i + 0 \cdot \sqrt{U^2 - \|x_i\|^2 - b_i^2} = \langle h, x_i \rangle + b_i$. The second equation follows from the relation between the inner product and the distance in Euclidean spaces: $\rho(\tilde{h}, \phi(x_i, b_i))^2 = \|\tilde{h} - \phi(x_i, b_i)\|^2 = \|\tilde{h}\|^2 + \|\phi(x_i, b_i)\|^2 - 2\langle \tilde{h}, \phi(x_i, b_i) \rangle = \|h\|^2 + 1 + U^2 - 2\langle \tilde{h}, \phi(x_i, b_i) \rangle$. □

*Theorem 3.2 proof.* By Lemma 3.1, the distance $\rho(\tilde{h}, \phi(x_i, b_i))$ depends strictly monotonically decreasingly on $\langle h, x_i \rangle + b_i$. The claim of the theorem then follows straightforwardly upon this fact. □

## B A Bound on Top-$K$ Softmax Approximation

In this section we derive an error bound of top-$K$ highest ranked words approximation on softmax toward a probability distribution. Let $V$ be the set of vocabulary. Let $s_i$ ($1 \leq i \leq |V|$) be the scores produced using exhaustive search, sorted in decreasing order. Suppose $L$ is a lower bound for $\{s_i\}$, i.e., $L \leq \min s_i$. Typically one chooses $L$ to be a negative number with sufficiently large absolution value. If there is no known lower bound, one may also set $L = -\infty$ and agrees that $\exp(-\infty) = 0$. The probability distribution generated by applying softmax on $s_i$ is given by $p_i = \exp(s_i) / \sum_i \exp(s_i)$. Using exhaustive search we are able to compute $p_i$ as exact. However with approximated techniques, we are only able to obtain an approximation $\hat{p}_i$ of the distribution. In real applications we typically only care about how $\hat{p}_i$ differs from $p_i$ for the top-$K$ words. The error in such approximations comes from two sources: (1) the accuracy of the approximation to obtain the approximated top-$K$; and (2) the approximation of $\sum_i \exp(s_i)$. In the following theorem we give a quantitative analysis of how large the relative error could be.

**Theorem B.1.** *Let $V$, $s_i$, $L$, $p_i$ be as above. Let $\mathcal{K} \subseteq \{1, \ldots, |V|\}$ be the ground truth top-$K$ indices. Suppose an approximation top-$K$ softmax algorithm gives $\mathcal{K}' \subseteq \{1, \ldots, |V|\}$ as the approximated top-$K$ indices. Let $\hat{s}_i$ be the approximated score the algorithm assigns to the $i$-th word in $V$, and let $s' = \min_{i \in \mathcal{K}'} \hat{s}_i$. Assume that (i) the algorithm assigns exact scores $\hat{s}_i = s_i$ to those $i \in \mathcal{K}'$; and (ii) it assigns a score $\hat{s}_i$ for $i \notin \mathcal{K}'$ such that $L \leq \hat{s}_i \leq s'$. The approximated probability distribution is given by $\hat{p}_i := \exp(\hat{s}_i) / \sum_i \exp(\hat{s}_i)$. Let $\mathcal{K}'' = \{i : s_i \geq s'\}$. Then the relative error of probability distribution approximation is bounded by $|\hat{p}_i - p_i| / p_i \leq$*
$$\frac{\sum_{i \in \mathcal{K}'' \setminus \mathcal{K}'} (\exp(s_i) - \exp(L)) + (|V| - |\mathcal{K}''|)(\exp(s') - \exp(L))}{\sum_{i \in \mathcal{K}'} \exp(\hat{s}_i) + (|V| - K) \exp(L)} \text{ for any } i \in \mathcal{K} \cap \mathcal{K}'.$$

*Proof.* First note that $p_i = \frac{\exp(s_i)}{\sum_i \exp(s_i)}$ and that $\hat{p}_i = \frac{\exp(\hat{s}_i)}{\sum_i \exp(\hat{s}_i)}$. Since $i \in \mathcal{K} \cap \mathcal{K}'$, we have $s_i = \hat{s}_i$. We then deduce that

$$\frac{|\hat{p}_i - p_i|}{p_i} = \frac{|\sum_i \exp(s_i) - \sum_i \exp(\hat{s}_i)|}{\sum_i \exp(\hat{s}_i)}. \tag{1}$$

We then proceed to bound both the numerator and the denominator.

To find an upper bound for the numerator, first note that

$$\left| \sum_i \exp(s_i) - \sum_i \exp(\hat{s}_i) \right| \leq \left| \sum_{i \in \mathcal{K}''} \exp(s_i) - \sum_{i \in \mathcal{K}''} \exp(\hat{s}_i) \right| + \left| \sum_{i \notin \mathcal{K}''} \exp(s_i) - \sum_{i \notin \mathcal{K}''} \exp(\hat{s}_i) \right|. \tag{2}$$

For the first summand, first observe that $\mathcal{K} \subseteq \mathcal{K}''$ and $\mathcal{K}' \subseteq \mathcal{K}''$. Therefore

$$\left| \sum_{i \in \mathcal{K}''} \exp(s_i) - \sum_{i \in \mathcal{K}''} \exp(\hat{s}_i) \right| = \left| \sum_{i \in \mathcal{K}'} (\exp(s_i) - \exp(\hat{s}_i)) + \sum_{i \in \mathcal{K}'' \setminus \mathcal{K}'} (\exp(s_i) - \exp(\hat{s}_i)) \right|.$$

The condition (i) implies that the first sum is zero. The second sum is always non-negative since $\exp(s_i) \geq \exp(s') \geq \exp(\hat{s}_i)$ for $i \in \mathcal{K}''\backslash\mathcal{K}'$. Thus $\left|\sum_{i\in\mathcal{K}''\backslash\mathcal{K}'}(\exp(s_i) - \exp(\hat{s}_i))\right| = \sum_{i\in\mathcal{K}''\backslash\mathcal{K}'}(\exp(s_i) - \exp(\hat{s}_i)) \leq \sum_{i\in\mathcal{K}''\backslash\mathcal{K}'}(\exp(s_i) - \exp(L))$. For the second summand in the right hand side of Equation (2), note that $\left|\sum_{i\notin\mathcal{K}''}\exp(s_i) - \sum_{i\notin\mathcal{K}''}\exp(\hat{s}_i)\right| \leq \sum_{i\notin\mathcal{K}''}|\exp(s_i) - \exp(\hat{s}_i)|$. Since both $s_i, \hat{s}_i \in [L, s']$ for all $i \notin \mathcal{K}''$, we have $\sum_{i\notin\mathcal{K}''}|\exp(s_i) - \exp(\hat{s}_i)| \leq \sum_{i\notin\mathcal{K}''}(\exp(s') - \exp(L)) = (|V| - |\mathcal{K}''|)(\exp(s') - \exp(L))$. This shows the upper bound of the numerator.

For the denominator in the right hand side of Equation (1), simply note that $\sum_i \exp(\hat{s}_i) = \sum_{i\in\mathcal{K}'}\exp(\hat{s}_i) + \sum_{i\notin\mathcal{K}'}\exp(\hat{s}_i) \geq \sum_{i\in\mathcal{K}'}\exp(\hat{s}_i) + (|V| - K)\exp(L)$. This then concludes the proof of the theorem. □

It is worthy to point out that the numerator of the above error bound can be rewritten as $\sum_{i\in\mathcal{K}''\backslash\mathcal{K}'}(\exp(s_i) - \exp(s')) + (|V| - K)(\exp(s') - \exp(L))$. Intuitively, the theorem states that the accuracy of the softmax probability approximation for the top-$K$ words depends on three quantities: (i) $\sum_{i\in\mathcal{K}''\backslash\mathcal{K}'}(\exp(s_i) - \exp(s'))$, which measures how many words are "missed" by the approximation of top-$K$ words. (ii) $\exp(s') - \exp(L)$, which measures the distribution of the scores found by the approximation. The smaller (i) and (ii) are (relative to $\sum_{i\in\mathcal{K}'}\exp(\hat{s}_i)$), the better the approximation is.

We also observe that when the precision at $K$ is 1 for the approximation algorithm, then the bound depends only on the sum of exponential scores and the smallest top-$K$ score retrieved by the algorithm.

**Corollary B.1.1.** *Let the notations be the same as in the above theorem. Assume that the precision at $K$ of the approximation is 1. Further assume that all the scores $s_i$ are distinct. Then the relative error to the approximated softmax probability distribution is bounded above by $|\hat{p}_i - p_i|/p_i \leq \dfrac{(|V| - K)(\exp(s') - \exp(L))}{\sum_{i\in\mathcal{K}'}\exp(\hat{s}_i) + (|V| - K)\exp(L)}$ for any $i \in \mathcal{K} \cap \mathcal{K}'$.*

*Proof.* It suffices to show that $\mathcal{K}'' = \mathcal{K}'$. Since the precision at $K$ is 1, we have $\mathcal{K} = \mathcal{K}'$, which means $s' = \min_{i\in\mathcal{K}'}\hat{s}_i = \min_{i\in\mathcal{K}}s_i$. Now assume $i \in \mathcal{K}''$, then by definition of $\mathcal{K}''$, $s_i \geq \min_{j\in\mathcal{K}}s_j$. Since all scores are distinct, this shows that $s_i$ is amongst the top-$K$, i.e., $s_i \in \mathcal{K} = \mathcal{K}'$. Thus $\mathcal{K}'' \subseteq \mathcal{K}'$. The other direction of inclusion is trivial. □

## C Additional Results

**Speedup with different vocabulary size.** Figure 5 shows the speedup for FGD (with different $efSearch$) over the execution time of full-softmax for vocabulary size 10K, 20K, 40K, and 80K. When $efSearch = 20$, FGD achieves more than 65X speedup over the baseline with a vocabulary size 80K. Even with smaller vocabulary size, FGD still achieves roughly an order of magnitude speedup. Overall, FGD achieves speedup over the baseline consistently and scales well with different $efSearch$ values.

**Sensitivity of word embedding dimension.** Table 4 reports the precision with varying word vector embedding dimension 128, 256, and 512 on the WikiText-2 language modeling. The vocabulary size is set to the default 33,728. In most cases, $efSearch$ being 50 or 100 is sufficient to provide high precision (e.g., > 0.95). Over 0.99 precision can be reached when $efSearch$ is 200. This indicates that FGD offers high precision with different word embedding dimensions.

Figure 6 compares with the execution time of FGD and full-softmax, FGD achieves an order of magnitude reduction of execution time.

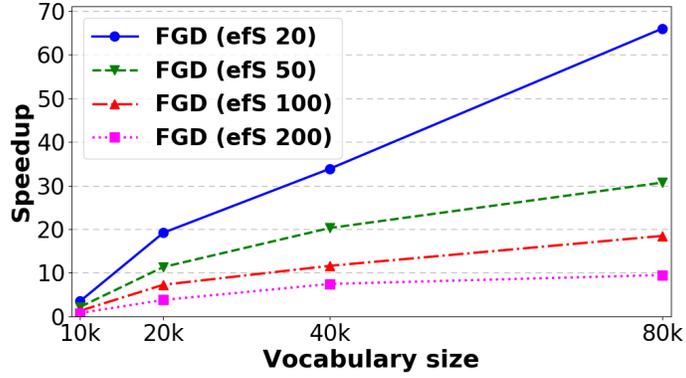

Figure 5: Performance of FGD, normalized to full-softmax execution time. Higher is better.

| D | P@K | FGD (efSearch) | | | |
|---|---|---|---|---|---|
| | | 20 | 50 | 100 | 200 |
| 128 | P@1 | 0.913 | 0.993 | 0.998 | 0.999 |
| | P@10 | 0.819 | 0.934 | 0.976 | 0.992 |
| 256 | P@1 | 0.832 | 0.917 | 0.958 | 0.992 |
| | P@10 | 0.866 | 0.944 | 0.976 | 0.995 |
| 512 | P@1 | 0.854 | 0.921 | 0.968 | 0.988 |
| | P@10 | 0.884 | 0.950 | 0.979 | 0.995 |

Table 4: Precision of FGD on WikiText-2 dataset varying word vector embedding dimension.

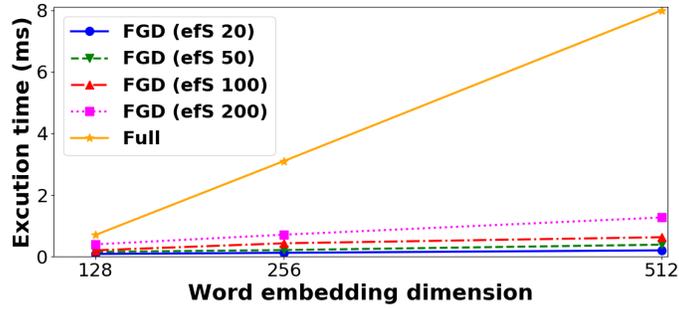

Figure 6: Scalability of WikiText-2 language model varying word vector embedding dimension.

6